\documentclass[10pt,twocolumn,letterpaper]{article}
\usepackage{iccv}
\usepackage[pdftex]{graphicx} 
\usepackage{times}
\usepackage{epsfig}
\usepackage{amsmath}
\usepackage{amssymb}
\usepackage{multirow}
\usepackage[numbers]{natbib}
\usepackage{paralist}
\usepackage{subcaption}
\usepackage{makecell}
\usepackage{mwe}

\DeclareMathOperator*{\argmin}{arg\,min}

\usepackage[pagebackref=true,breaklinks=true,letterpaper=true,colorlinks,bookmarks=false]{hyperref}

\iccvfinalcopy 
\ificcvfinal\fi

\usepackage{subcaption}
\usepackage{amsmath} 
\usepackage{amssymb}  
\usepackage{csquotes}

\usepackage{array}
\usepackage{enumerate}
\usepackage{multirow} 
\usepackage{float}
\usepackage{paralist}
\usepackage{amsfonts}
\usepackage{longtable}

\newcolumntype{L}[1]{>{\raggedright\let\newline\\\arraybackslash\hspace{0pt}}m{#1}}
\newcolumntype{C}[1]{>{\centering\let\newline\\\arraybackslash\hspace{0pt}}m{#1}}
\newcolumntype{R}[1]{>{\raggedleft\let\newline\\\arraybackslash\hspace{0pt}}m{#1}}

\graphicspath{ {img/} }

\newcommand{\nostarnote}[1]{}
\newcommand{\baad}[1]{} 
\usepackage[size=small]{caption}
\usepackage{color}

\usepackage{makecell}

\usepackage{color}

\begin{document}
\title{UDepth: Fast Monocular Depth Estimation for Visually-guided Underwater Robots}  

\author{Boxiao Yu, Jiayi Wu and Md Jahidul Islam \\
{\tt\small \{boxiao.yu@, wuj2@, jahid@ece.\}ufl.edu} \\
{
\small RoboPI Laboratory, Dept. of Electrical and Computer Engineering (ECE)} \\ 
{
\small University of Florida (UF), Gainesville, FL, USA }
\thanks{This pre-print is accepted for publication at the ICRA 2023. [01/23]}
}

\maketitle
\thispagestyle{empty}

\begin{abstract}
\textit{In this paper, we present a fast monocular depth estimation method for enabling 3D perception capabilities of low-cost underwater robots. We formulate a novel end-to-end deep visual learning pipeline named UDepth, which incorporates domain knowledge of image formation characteristics of natural underwater scenes. First, we adapt a new input space from raw RGB image space by exploiting underwater light attenuation prior, and then devise a least-squared formulation for coarse pixel-wise depth prediction. Subsequently, we extend this into a domain projection loss that guides the end-to-end learning of UDepth on over 9K RGB-D training samples. UDepth is designed with a computationally light MobileNetV2 backbone and a Transformer-based optimizer for ensuring fast inference rates on embedded systems. By domain-aware design choices and through comprehensive experimental analyses, we demonstrate that it is possible to achieve state-of-the-art depth estimation performance while ensuring a small computational footprint. Specifically, with 70\%-80\% less network parameters than existing benchmarks, UDepth achieves comparable and often better depth estimation performance. While the full model offers over 66 FPS (13 FPS) inference rates on a single GPU (CPU core), our domain projection for coarse depth prediction runs at 51.5 FPS rates on single-board NVIDIA\texttrademark~Jetson TX2s. The inference pipelines are available at \url{https://github.com/uf-robopi/UDepth}}.
\end{abstract}

\section{Introduction}

\begin{figure}[t]
\centering
\includegraphics[width=\linewidth]{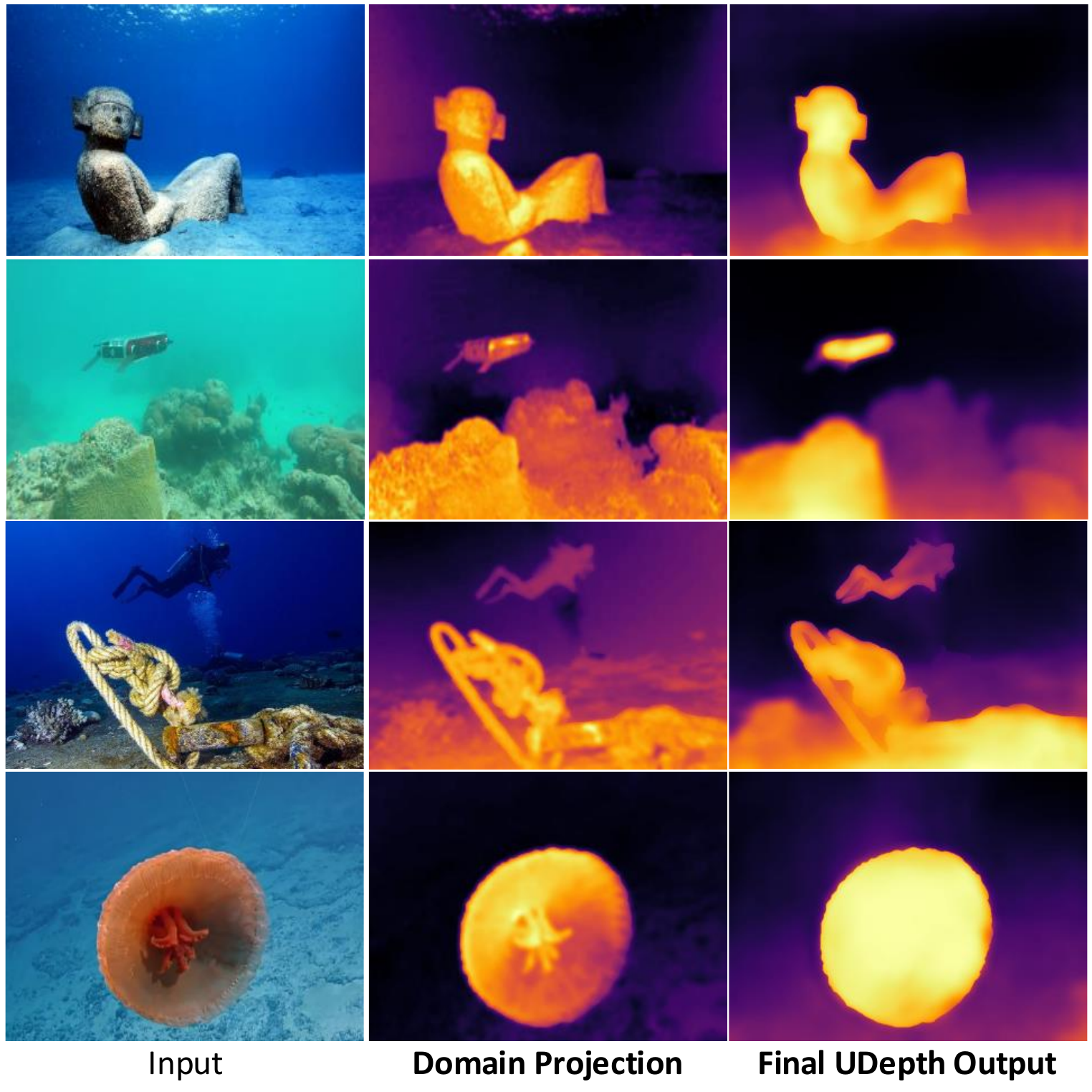}
\vspace{-5mm}
\caption{The proposed UDepth learning pipeline uses the wavelength-dependent attenuation constraints of underwater domain for an intelligent input-space adaptation from raw RGB images. A computationally light domain projection step then approximates a coarse depth map. The intermediate projections and other pixel-level loss functions guide the end-to-end training of UDepth for fine-grained depth estimations.}
\label{fig:intro}
\vspace{-3mm}
\end{figure}

Underwater depth estimation is the foundation for numerous marine robotics tasks such as autonomous mapping and 3D reconstruction~\cite{digumarti2016underwater}, visual filtering~\cite{islam2020fast}, tracking and servoing~\cite{9340919,raanan2018detection}, navigation~\cite{xanthidis2020navigation,rahman2019svin2}, photometry and imaging technologies~\cite{akkaynak2018revised}, and more. Unlike terrestrial robots, visually guided underwater robots have very few low-cost solutions for dense 3D visual sensing because of the high cost and domain-specific operational complexities involved in deploying underwater LiDARs~\cite{zhou2021overview,mcleod2013autonomous}, RGB-D cameras~\cite{7891971}, or laser scanners~\cite{palomer2019underwater}. Fast monocular depth estimation thus plays a critical role in enabling real-time 3D robot perception and state estimation for important applications such as subsea monitoring and inspection~\cite{shkurti2012multi}, autonomous exploration~\cite{girdhar2014autonomous}, and companion robotics~\cite{islam2021robot}. 

Traditional approaches for underwater depth estimation either use active photon counting on Time-of-flight (ToF) cameras~\cite{maccarone2015underwater} or geometric model adaptation on structure light~\cite{massot2015optical,8559463}. Standard depth cameras~\cite{7891971} with water-housing and echo-sounders~\cite{9340919} are also common to acquire underwater scene depths up-to-scale. However, accurate sensory integration and backscatter filtering are major challenges in their field deployments. Another class of approaches holistically evaluates the physical properties of the optics for depth estimation; \ie, they approximate the medium transmission rate, background light, and attenuation coefficient following an underwater image formation model (IFM)~\cite{8084744,7351749,raihan2020depth,chang2018single}, and subsequently estimate depth for each pixel. These methods work reasonably well on clear/non-turbid water-bodies (\eg, Ocean Type I-II)~\cite{song2018rapid}, and are particularly suitable for offline image processing tasks. However, their generalized end-to-end implementations are rather difficult due to the optical parameters' dependency on wavelength and water-body properties~\cite{akkaynak2018revised}.  

In recent years, deep visual learning-based methods are used to address these problems with remarkable success~\cite{hambarde2021uw,ye2019deep}. These data-driven methods heavily rely on large-scale datasets, hence researchers often synthetically generate training images~\cite{zhao2021synthesis,zhao2021unpaired,cui2021underwater} due to the lack of RGB-D ground truths for underwater scenes. However, such training pipelines cannot fully capture the complex underwater scene geometry and image statistics, restricting their performance and applicability for practical applications. Additionally, their learning objectives are generally adopted from standard terrestrial depth estimation pipelines or coupled with in-air RGB-D image pairs~\cite{ye2019deep,gupta2019unsupervised}. Consequently, underwater domain knowledge is not incorporated into their learning processes, which leads to poor generalization performance.  

In this paper, we formulate a robust and efficient end-to-end model named \textbf{UDepth}, for fast monocular depth estimation by incorporating underwater domain knowledge into its supervised learning pipeline. We adapt a new input space R\textbf{M}I for underwater domain-aware learning, which consists of red channel (R), maximum of green and blue channels (\textbf{M}), and grayscale intensities (I). Due to the wavelength-dependent attenuation constraints of underwater light propagation, the relative differences between R and \textbf{M} channels embed useful scene depth information while the I channel preserves structural contents of the image. Based on R\textbf{M}I, we design a computationally light projection step for pixel-wise \textbf{coarse depth prediction} by a least-squared formulation. We further devise a \textbf{domain projection loss} to enforce the pixel-wise underwater attenuation constraints in the holistic learning process as well. 

The proposed UDepth model architecture consists of the highly efficient MobilenetV2 backbone, a lightweight Vision Transformer (mViT)-based global attention module, and a convolutional regressor for fine-grained depth estimation. The end-to-end inference graph of UDepth has only $15.6$M parameters, which is about $80\%$ less than Adabins~\cite{bhat2021adabins} and $66\%$ less than DenseDepth~\cite{alhashim2018high} - two of its closest competitor baselines. As a result, UDepth offers significantly faster inference rates: over $66.7$ FPS on a NVIDIA\texttrademark~RTX $3080$ and $13.3$ FPS on an Intel\texttrademark~Core i9-3.50GHz CPU core. With comprehensive domain-aware learning on $9223$ RGB-D pairs of natural underwater schenes from USOD10K datset~\cite{USOD10K}, UDepth achieves SOTA performance on standard benchmarks despite having such a light architecture. Moreover, we demonstrate that UDepth offers better generalization performance on arbitrary test cases from Sea-Thru dataset~\cite{akkaynak2019sea} and real-world field experimental data. 

Furthermore, we demonstrate that our domain projection module with subsequent filtering can be used for coarse depth prediction on low-power embedded devices. Specifically, it runs at $51.5$ FPS on NVIDIA\texttrademark~Jetson TX2s and $7.92$ FPS on Raspberry Pi-4s. More importantly, visual results on field experimental data suggest that its coarse predictions are reasonably accurate and can be used for on-board 3D perception by low-cost underwater robots.

\section{Background \& Related Work}
\vspace{-1mm}
\subsection{Monocular Depth Estimation Literature}
\vspace{-1mm}
Traditional methods for monocular depth estimation~\cite{godard2019digging,godard2017unsupervised} focus on graphical models with hand-crafted geometric priors based on the concepts of structure-from-motion, stereo vision, and multi-view feature matching. These classical methods require multi-view correspondences and significant computational power, yet only generate sparse depth information. Such difficulties have paved the way for powerful deep visual learning-based methods~\cite{zhao2020monocular}, which can learn to infer dense depth maps from single RGB images in an end-to-end manner. The state-of-the-art (SOTA) methods of the modern era for monocular depth estimation can be classified as supervised, semi-supervised, and unsupervised approaches.

\vspace{-3mm}
\subsubsection{Supervised Methods}
\vspace{-2mm}
With large-scale paired datasets generated by RGBD cameras, supervised training pipelines of fully-convolutional networks (FCNs) have dominated the SOTA performance over the past decade~\cite{zhao2020monocular}. The prototypical model architectures adopt progressively upsampled feature representations~\cite{fu2018deep,ranftl2020towards}, dilated convolutions~\cite{qiao2021vip}, or parallel multi-scale feature aggregation~\cite{lee2018single,li2017two} to learn fine-grained depth predictions from RGB images. More recent architectures integrate high-resolution representation with multiple lower-resolution \textit{refined} feature maps~\cite{eigen2014depth,eigen2015predicting} to improve the local structural details of the prediction. In recent years, various Vision Transformer (ViT)~\cite{vaswani2017attention}-based spatial \textit{attention blocks}~\cite{bhat2021adabins,ranftl2021vision} are incorporated to ensure global context awareness. Notable objective functions used by these networks are the reverse Huber loss~\cite{zwald2012berhu}, classification and ordinal regression loss~\cite{fu2018deep,cao2017estimating}, pairwise ranking loss~\cite{xian2020structure}, and other adaptive weighting or density losses~\cite{bhat2021adabins,lee2020multi}.

\vspace{-3.5mm}
\subsubsection{Semi-supervised and Unsupervised Methods}
\vspace{-1.5mm}
The unsupervised learning methods avoid the need for large-scale paired data by inferring scene depth from two or more RGB images by exploiting their inherent geometric constraints~\cite{9803821}. Contemporary methods take stereo image pairs~\cite{garg2016unsupervised, godard2017unsupervised} or consecutive frames from video~\cite{zhou2017unsupervised, godard2019digging} as inputs, combined with geometric view reconstruction~\cite{watson2019self, poggi2018learning} or camera pose estimation~\cite{guizilini20203d} to achieve dense monocular depth prediction. In recent years, attention mechanisms are coupled into such networks to preserve the inherent spatial details of the predicted depth map~\cite{dai2019unsupervised,xu2021multi}. Practical loss functions such as left-right disparity consistency loss~\cite{godard2017unsupervised, yang2018deep}, photometric loss~\cite{zhan2018unsupervised}, and symmetry loss~\cite{zhou2019unsupervised} are generally utilized in these networks as learning objectives. On the other hand, to get more scale information while avoiding costly ground truth, researchers proposed semi-supervised learning methods by combining multiple training stages~\cite{luo2018single, guo2018learning} or training loss terms~\cite{kuznietsov2017semi} of supervised and unsupervised learning. The left-right consistency loss~\cite{amiri2019semi}, mutual distillation loss~\cite{baek2022semi}, and teacher-student learning strategies~\cite{cho2019large} are also introduced into semi-supervised learning pipelines with inspiring results for robust monocular depth estimation.

\subsection{Depth Estimation on Underwater Imagery}
SOTA monocular depth estimation methods are not directly applicable off-the-shelf to underwater imagery due to their domain-specific image formation characteristics and scene geometry~\cite{akkaynak2018revised,drews2016underwater}. Researchers have addressed this in many ways, which can be categorized into active or passive approaches. Prominent \textit{active} approaches use time-correlated single-photon counting on ToF cameras~\cite{maccarone2015underwater} or 3D model adaptation on structure light~\cite{massot2015optical,8559463}. On the other hand, \textit{passive} estimation methods incorporate domain knowledge either following a physics-based IFM or from large-scale data; these models are more popular for their operational simplicity. While atmospheric models were used by early methods~\cite{8084665}, contemporary approaches generally adopt the Jaffe-McGlamery underwater IFM~\cite{6720385} to estimate the depth map from the medium transmission map, \ie, the fraction of scene radiance that reaches the camera after absorption and scattering. 

The medium transmission map is generally estimated by using the concepts of dark channel prior (DCP)~\cite{he2010single}, often adapted for underwater scenes as UDCP (underwater DCP)~\cite{7426236}, red-channel compensation~\cite{galdran2015automatic}, etc. Blurriness and illumination information are also used to approximate the transmission map and background light first, and then depth maps are estimated by following the IFM~\cite{8084744,7351749,raihan2020depth,chang2018single}. Hence, depth map estimation from the Jaffe-McGlamery model or following the (more comprehensive) revised IFM~\cite{akkaynak2018revised,akkaynak2019sea} requires accurate estimation of background light, transmission map, and water-body parameters. However, these parameters are not always known; in fact, they are estimated by using noisy depth priors for image enhancement and color correction~\cite{ancuti2017color,skinner2019uwstereonet}.

A practical alternative is to apply transfer learning on SOTA deep visual models for fast monocular depth estimation. The idea is to take advantage of large-scale terrestrial RGB-D datasets for pre-training, and then tune the model weights on limited underwater data. The most commonly adopted models are the Monodepth2~\cite{godard2019digging}, AdaBins~\cite{bhat2021adabins}, various U-Net~\cite{ronneberger2015u} based architectures (\eg, with VGG encoder~\cite{simonyan2014very,laina2016deeper} or ResNet50 encoder~\cite{he2016deep}), and GANs~\cite{hambarde2021uw}. Contemporary researchers have also proposed hybrid training pipelines where in-air (\ie, terrestrial) RGB-D pairs drive the supervised training, while the domain (\ie underwater) data is used in a self-supervised manner simultaneously~\cite{ye2019deep,gupta2019unsupervised}. Despite some early success of these existing approaches, several important aspects of ($i$) incorporating useful domain knowledge into comprehensive training pipelines, ($ii$) learning depth prediction from large-scale underwater RGB-D data, and ($iii$) computational feasibility analysis for robot vision - are not explored in the literature.

\begin{figure*}[t]
\centering
\includegraphics[width=0.55\linewidth]{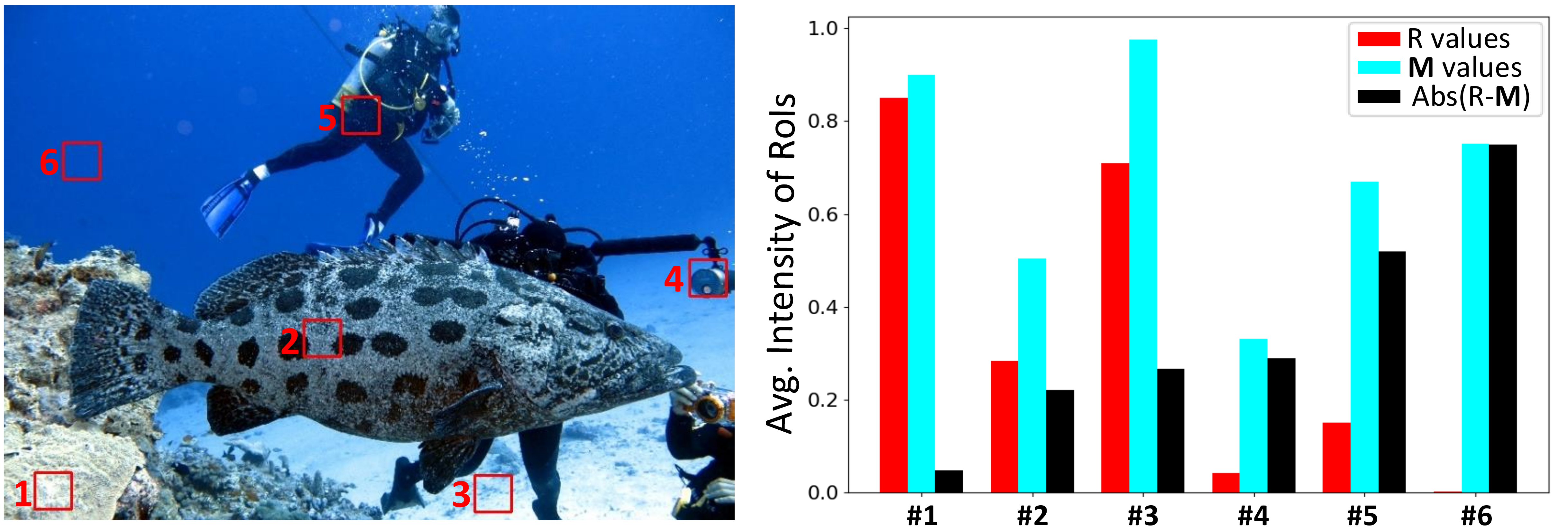}~
\includegraphics[width=0.38\linewidth]{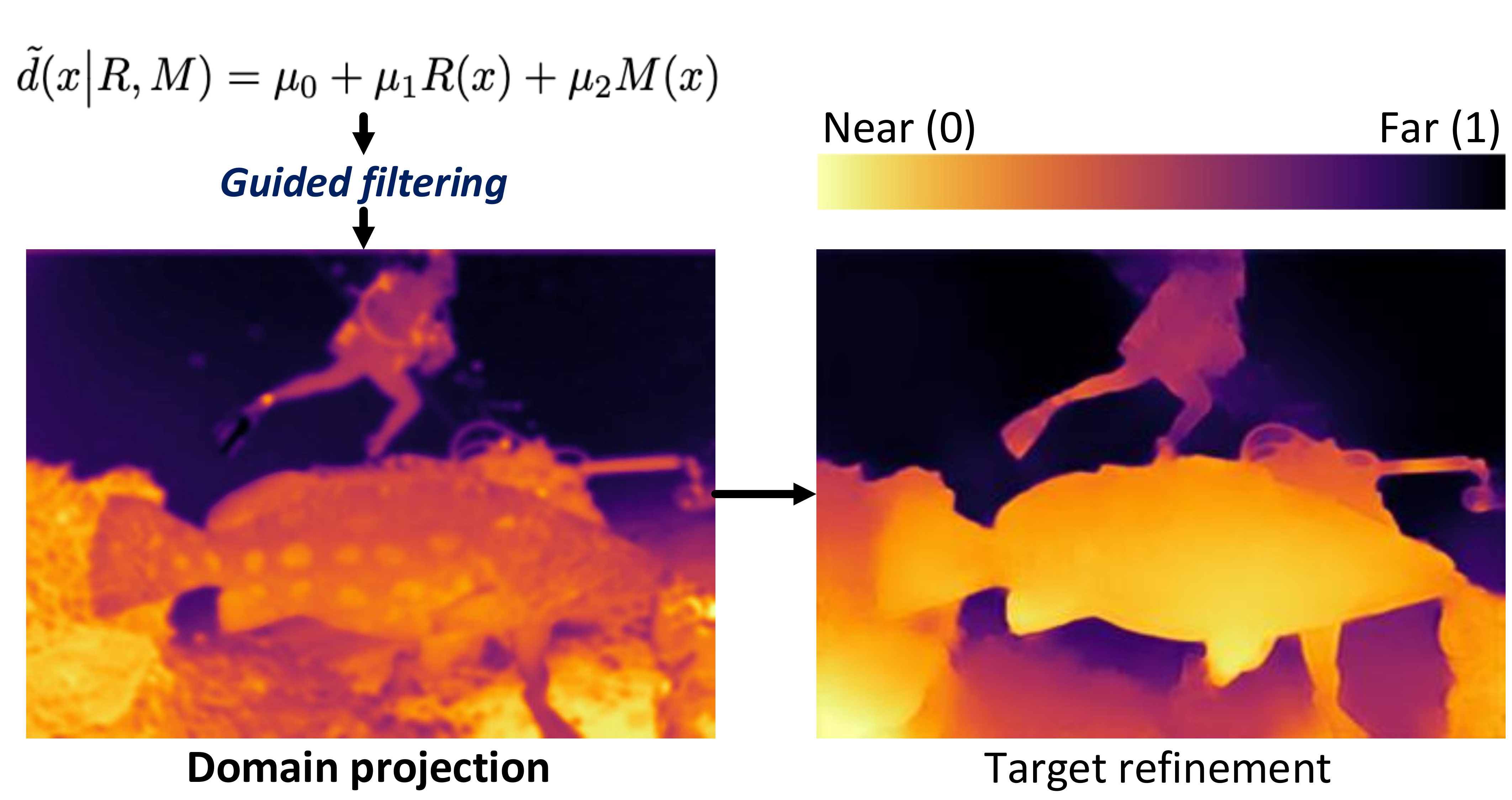}
\vspace{-2mm}
\caption{\small {Red light has the largest wavelength and thus gets attenuated the most underwater~\cite{song2018rapid,7426236}. Differences in R and \textbf{M}$=$$max$\{G,B\} values vary proportionately with pixel distances, as demonstrated by six $20\times20$ RoIs selected on the left image. We exploit this domain-specific attenuation constraint with a linear estimator for coarse depth predictions, which can be further filtered for smoothing; an example is shown on the right. Such abstract projection is the basis of our domain projection loss that guides UDepth learning.}}
\label{fig:rmg_space}
\end{figure*}

\section{Proposed Learning Pipeline}
\vspace{-1mm}
\subsection{New Input Space Adaptation: RMI}\label{sec_rmi}
\vspace{-1mm}
Although the wavelength-dependency of atmospheric light attenuation and scattering are negligible, they impact underwater image formation significantly. According to the Jaffe-McGlamery underwater IMF~\cite{6720385}, an observed image $\mathbf U_\lambda(\mathbf x)$ can be expressed as:
\begin{equation}
    \mathbf U_\mathbf \lambda(\mathbf x) = {\mathbf I_\mathbf \lambda(\mathbf x)\cdot t_\mathbf \lambda(\mathbf x)} + {\mathbf B_\mathbf \lambda \cdot (1- t_\mathbf \lambda(\mathbf x))}, 
    \label{eq4}
\end{equation}
where $\mathbf \lambda \in\{\text{R,G,B}\}$ is the wavelength component, $\mathbf I_\lambda(\mathbf x)$ is the clear latent image, $\mathbf B_\lambda$ denotes the global background light, and $t_\lambda(\mathbf x)$ represents the medium transmission rate. The $t_\mathbf \lambda(\mathbf x)$ is a function of pixel depth $d(x)$ and medium attenuation coefficient $\beta_\mathbf \lambda$, defined as: $t_\mathbf \lambda(\mathbf x) = e^{-\beta_\lambda \cdot d(\mathbf x)}$.

Many contemporary works use priors like DCP~\cite{he2010single} or UDCP~\cite{7426236} to approximate $t_\mathbf \lambda(\mathbf x)$ and estimate coarse scene depth $d(\mathbf x)$. Since red wavelength suffers more aggressive attenuation underwater, techniques such as underwater light attenuation prior (ULAP)~\cite{song2018rapid} and red-channel compensation~\cite{galdran2015automatic} can exploit the R channel values to further refine the depth prediction. As illustrated in Fig.~\ref{fig:rmg_space}, the relative differences between \{R\} and \{G, B\} channel values encode useful depth information for a given pixel. In this paper, we exploit these inherent relationships and demonstrate that R\textbf{M}I$\equiv$\{R, \textbf{M}$=$$max$\{G,B\}, I (intensity)\} is a significantly better input space for visual learning pipelines of underwater monocular depth estimation models.

\vspace{-0.5mm}
\subsection{Network Architecture: UDepth Model}
\begin{figure*}[t]
\centering    \includegraphics[width=0.95\linewidth]{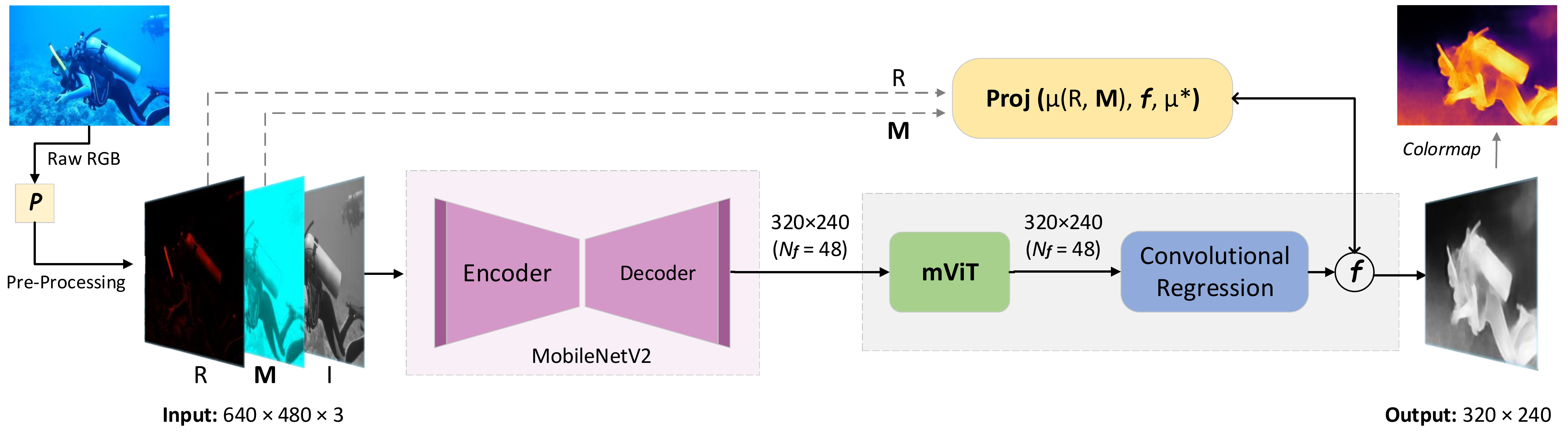}
\vspace{-3mm}
\caption{The end-to-end learning pipeline of our proposed UDepth model is shown. Raw RGB images are first pre-processed to map into R\textbf{M}I input space, then forwarded to the MobileNetV2 backbone for feature extraction. Those features are refined by a transformer-based optimizer (mViT), followed by a convolutional regressor to generate the single channel depth prediction. The learning objective involving pixel-wise losses and a domain projection loss is formulated in Eq.~\ref{obj_fun}.}
\label{fig:udepth_arch}
\vspace{-5mm}
\end{figure*}

\vspace{-1mm}
As illustrated in Fig.~\ref{fig:udepth_arch}, the network architecture of UDepth model consists of three major components: a MobileNetV2-based encoder-decoder backbone, a transformer-based refinement module (mViT), and a convolutional regressor. These components are tied sequentially for the supervised learning of monocular depth estimation.

\vspace{-3mm}
\subsubsection{\textbf{MobileNetV2 backbone}} 
\vspace{-1mm}
We use an encoder-decoder backbone based on MobileNetV2~\cite{sandler2018mobilenetv2} as it is highly efficient and designed for resource-constrained platforms. It is considerably faster than other SOTA alternatives with only a slight compromise in performance, which makes it feasible for robot deployments. It is based on an \textit{inverted residual} structure with residual connections between \textit{bottleneck} layers~\cite{sandler2018mobilenetv2}. The intermediate expansion layers use lightweight depthwise convolutions to filter features as a source of non-linearity. The encoder contains a series of fully convolution layers with $32$ filters, followed by a total of $19$ residual bottleneck layers. We adapt the last convolutional layer of decoder so that it finally generates $48$ filters of $320\times 480$ resolution, given a 3-channel R\textbf{M}I input.
\vspace{-3mm}
\subsubsection{\textbf{mViT refinement}}
\vspace{-1mm}
Transformers can perform global statistical analysis on images, solving the problem that traditional convolution models can only handle pixel-level information~\cite{dosovitskiy2020image}. Due to the heavy computational cost of Vision Transformers (ViT), we adopt a lighter mViT architecture inspired by~\cite{bhat2021adabins}. The $48$ filters extracted by the backbone are $1\times1$ convolved and flattened to patch embeddings, which serve as inputs to the mViT encoder. Those are also fed to a $3\times3$ convolutional layer for spatial refinements. The $1 \times 1$ convolutional kernels are subsequently exploited to compute the range-attention maps $\mathbf R$, which combines adaptive global information with local pixel-level information from CNN. The other embedding is propagated to a multilayer perceptron head with ReLU activation to obtain a $80$-dimensional \textit{bin-width} feature vector $\mathbf f_b$. 

\vspace{-3mm}
\subsubsection{\textbf{Convolutional regression}} 
\vspace{-1mm}
Finally, the convolutional regression module combines the range-attention maps $\mathbf R$ and features $\mathbf f_b$ to generate the final feature map $\mathbf f$. To avoid discretization of depth values, the final prediction of depth map $\mathbf D$ is computed by the linear combination of bin-width centers ($\bar{\mathbf f_{b}}$), which is given by: $\hat{d} = \sum_{k} \bar{\mathbf f_{bk}} \; \sigma(\mathbf R_k)$.

\subsection{Objective Function Formulation}
\noindent
{\textbf{Pixel-wise depth losses.}} We use two pixel-wise supervised loss functions: $i)$ the $\mathcal{L}_2$ loss, and $ii)$ a scaled version of the Scale-Invariant Log (SILog) loss introduced by Eigen \etal~\cite{eigen2014depth}. These are defined as follows:
\begin{align}
\footnotesize
    \mathcal{L}_{2} &= \mathbb{E}\Big[ \big\lVert {d}_{i} - \hat{d}_{i} \big\rVert_2 \Big],  \\
    \mathcal{L}_{SILog}&=\alpha\sqrt{\frac{1}{T}\sum_{i}{g_i^2-\frac{\lambda}{T^2}(\sum_{i}{g_i})^2}} .
\end{align}
Here, $g_i=\log_{}{\hat{d_i}}-\log_{}{d_i}$ where $\hat{d}$ is the predicted depth, $d$ is the ground truth depth, and $T$ denotes the number of pixels having valid ground truth values. Inspired by~\cite{bhat2021adabins}, we use $\lambda=0.85$ and $\alpha=10$ in our implementation.
\vspace{1mm}

\noindent
{\textbf{Domain projection loss.}\label{sec_dpj}} We formulate a novel domain projection loss function based on our input space calibration. Following our discussion in Sec.~\ref{sec_rmi} and Fig.~\ref{fig:rmg_space}, we express the R-\textbf{M} relationship with the depth of a pixel $x$ with a linear approximator as follows: 
\begin{equation}
 \tilde{d}(x \big| R, M) = \mu_0 + \mu_1 R(x) + \mu_2 M(x).
\label{ulap_constraint}
\end{equation}
Then, we find the least-squared solution $\mu^*$ on the entire RGB-D training pairs, which is over $2.8$ billion pixels ($9229$ images of $640\times480$ resolution) by optimizing:
\begin{equation}
 \mu^* = \argmin_\mu \sum_{i,x} {\big\lVert {d}_{i}(x) - \tilde{d}(x \big| R_i, M_i) \big\rVert}_2^2.
\label{rmi_opt}
\end{equation}
We find $\mu^*=[0.496, -0.389, 0.464]$ in our experiments on USOD10K dataset~\cite{USOD10K} (see Sec.~\ref{usod_data}). Here, our goal is to use the optimal $\mu$-space for regularization, we penalize any pixel-wise depth predictions that violate the underwater image attenuation constraint defined by Eq.~\ref{ulap_constraint}. We achieve this by the following projection error function: 
\begin{equation}
  \mathcal{L}_{\perp} = \mathbb{E}\Big[\Pi_{\mu^*}\big(\tilde{d}_i(R, M) - \hat{d}_i) \big) \Big]
    \label{perp}
\end{equation}

\noindent
{\textbf{End-to-end objective}.} Finally, the end-to-end learning objective of our proposed UDepth pipeline is formulated as:  
\begin{equation}
 \mathcal{L}_{UDepth} = \lambda_2 \; \mathcal{L}_{2} + \lambda_{S}  \;\mathcal{L}_{SILog} + \lambda_{\perp} \; \mathcal{L}_{\perp}.
\label{obj_fun}
\end{equation}
In Eq.~\ref{obj_fun}, we find the $\lambda$-parameters empirically through hyper-parameters tuning. The optimal values used in UDepth training are: $\lambda_2 = 0.3$, $\lambda_{S} = 0.6$, and $\lambda_{\perp} = 0.1$.
\begin{figure*}[t]
\centering
\includegraphics[width=0.98\linewidth]{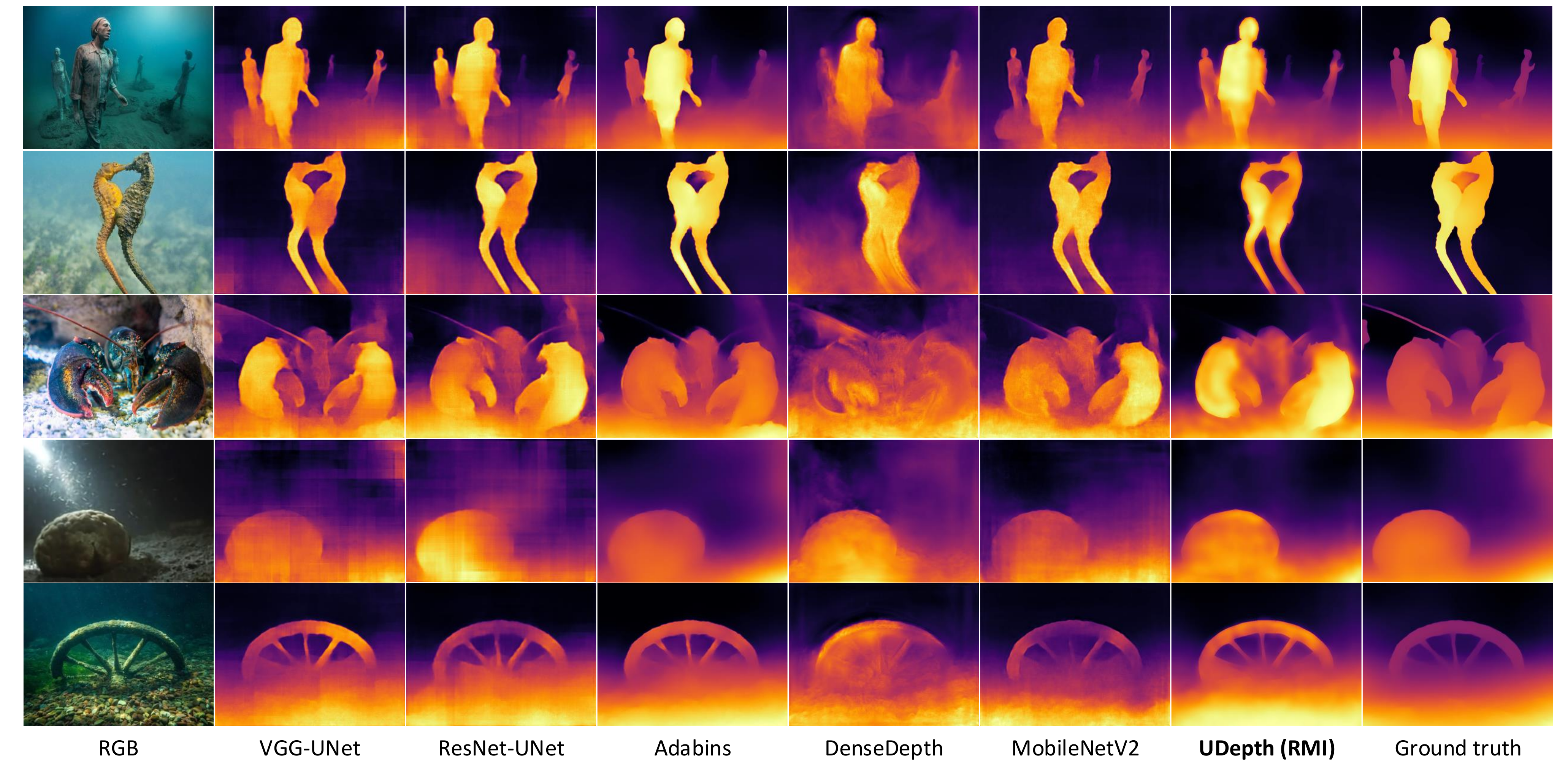}
\vspace{-2mm}
\caption{A few qualitative comparisons are shown for underwater scene depth estimation by UDepth and SOTA monocular depth estimation models on USOD10K test set~\cite{USOD10K}. As seen, UDepth infers accurate and consistent depth predictions across various waterbody, attenuation levels, and lighting conditions. UDepth offers comparable and often better performance than its closest competitor baseline Adabins, while offering $5\times$ faster inference and memory efficiency. 
}
\label{fig:qual_comp}
\vspace{-5mm}
\end{figure*}

\section{Experiments}
\subsection{Datasets and Evaluation Metrics}\label{usod_data}
\subsubsection{\textbf{Training and evaluation data}} 
\vspace{-1mm}
We use the USOD10K dataset~\cite{USOD10K} in our experiments; it contains RGB images and ground truth depth maps for various underwater scenes captured at a pixel resolution of $640\times480$. The dataset contains $9229$ training samples and $1026$ testing samples. We also use benchmark images from the Sea-Thru dataset~\cite{akkaynak2019sea} for performance evaluation. Moreover, we test UDepth model on unseen field data collected during oceanic explorations and human-robot collaborative experiments. 

\vspace{-3mm}
\subsubsection{\textbf{Evaluation metrics}} 
\vspace{-1mm}
We use four standard metrics~\cite{eigen2014depth} to compare our method against other SOTA models. These error  metrics are defined as follows: 
\begin{itemize}
    \item Mean absolute relative error (Abs Rel): $ \frac{1}{n} \sum_p^n \frac{|d_p - \hat{d}_p|}{d}$,
    \item Squared relative error (Sq Rel): $ \frac{1}{n} \sum_d^n \frac{||d_p - \hat{d}_p||^2}{d}$,
    \item Root mean squared error (RMSE): $\sqrt{\frac{1}{n}\sum_d^n(d_p - \hat{d}_p)^2}$, \item $\log_{10}$ error: $\frac{1}{n} \sum_d^n |\log_{10}{(d_p)}-\log_{10}{(\hat{d}_p)}|$,
\end{itemize}
Where $d_p$ is a pixel in depth image $d$, $\hat{d_p}$ is a pixel in the predicted depth image $\hat{d}$, and $n$ is the total number of pixels in $d$ (as well as $\hat{d}$). 

\subsection{Implementation Details}
UDepth training is supervised by RGB-D image pairs of natural underwater scenes; the RGB color images are pre-processed to prepare R\textbf{M}I input images and the respective D channel depth maps serve as ground truth. A total of $7383$ images in the USOD10K training dataset are used for training and the remaining $1846$ images are used for validation. We use Pytorch libraries~\cite{paszke2019pytorch} to implement its learning pipeline; AdamW~\cite{loshchilov2017decoupled} is used as the optimizer with an initial learning rate of $0.0001$ and exponential decay adjustment with a multiplicative factor of $0.9$. With a batch size of $4$, Udepth training takes about $5$ minutes per epoch on a single node with NVIDIA\texttrademark~RTX $3080$ GPU. It has about $15.5$M parameters in total: $3.5$M for the MobileNetV2 encoder, $10.9$M for the decoder, and $1.1$M for the mViT-based refiner and the convolutional regressor combined. For visual image generation, we apply guided filtering~\cite{he2012guided} to the raw UDepth output for smoothing by using binary saliency maps as the guided mask. The saliency masks can be generated by using traditional Salient Object Detection (SOD) models~\cite{Wu_2019_CVPR} if they are not available.



\subsection{Qualitative and Quantitative Evaluation}
For baseline performance comparison, we consider the following five models that are widely used for supervised learning of monocular depth estimation: VGG-UNet~\cite{simonyan2014very, ronneberger2015u}, ResNet-UNet~\cite{he2016deep, ronneberger2015u}, Adabins~\cite{bhat2021adabins}, DesneDepth~\cite{alhashim2018high},  and MobileNetV2~\cite{sandler2018mobilenetv2}. We train these models by following their recommended settings on the same pipeline and \textit{train-validation} data splits as UDepth. The qualitative performances of all these models for some samples are illustrated in Fig.~\ref{fig:qual_comp}, while the quantitative results are listed in Table~\ref{tab_usod}-\ref{tab_seathru}.

The two major findings of our experimental analyses are as follows: ($i$) All models exhibit consistently better results across almost all metrics when trained on the R\textbf{M}I space instead of raw RGB inputs, which validates our contribution to domain-aware input space adaptation. ($ii$) The proposed UDepth model outperforms VGG-UNet, ResNet-UNet, DesneDepth, and MobileNetV2 regardless of using R\textbf{M}I or RGB input space. On some metrics, the SOTA model Adabins achieves better scores and generate more accurate visual results. However, Adabins is a significantly heavier model with $78$M+ parameters compared to UDepth, which has only $15.5$M parameters. Hence, our design choices enable UDepth to achieve comparable and often better depth estimation performance than Adabins, despite being over $5\times$ more efficient (at only $20\%$ computational cost).



\begin{table}[t]
\centering
\caption{Quantitative comparison for underwater scene depth estimation performance by UDepth and other SOTA models on USOD10K test set~\cite{USOD10K}. Here, lower scores represent better performance for all metrics in consideration.}
\footnotesize
\begin{tabular}{l||l|l|l|l}
  \Xhline{2\arrayrulewidth}
   \textbf{Model} & Abs Rel & Sq Rel & RMSE & $\log_{10}$ \\ 
    
    \Xhline{2\arrayrulewidth} 
    VGG-UNet (RGB) & $0.939$ & $0.183$ & $0.150$ & $0.199$ \\
    
    VGG-UNet (R\textbf{M}I) & $0.886$ & $0.180$ & $0.150$ & $0.197$  \\
    
    ResNet50-UNet (RGB) & $0.875$ & $0.171$ & $0.152$ & $0.200$  \\
    
    ResNet50-UNet (R\textbf{M}I) & $0.847$ & $0.164$ & $0.154$ & $0.204$ \\
    
    AdaBins (RGB) & {$0.617$} & {$0.097$} & {$0.109$} & {$0.154$} \\
    
    AdaBins (R\textbf{M}I) & {{$0.613$}} & {{$0.090$}} & {{$0.108$}} & {{$0.153$}} \\
    
    
    DenseDepth (RGB) & $1.178$ & $0.254$ & $0.168$ & $0.219$ \\

    DenseDepth (R\textbf{M}I) & $1.128$ & $0.241$ & $0.170$ & $0.220$ \\
    
    MobileNetV2 (RGB) & $0.858$ & $0.156$ & $0.134$ & $0.183$ \\
    
    MobileNetV2 (R\textbf{M}I) & $0.790$ & $0.144$ & $0.136$ & $0.184$ \\
    
    UDepth (RGB) & $0.809$ & $0.147$ & {$0.129$} & {$0.176$}  \\
    
    
    UDepth (R\textbf{M}I) & {$0.681$} & {$0.123$} & $0.143$ & $0.188$ \\
    \Xhline{2\arrayrulewidth} 
\end{tabular}
\label{tab_usod}
\vspace{-5mm}
\end{table}

\begin{table}[t]
\centering
\caption{Quantitative performance comparison on Sea-thru (D3) dataset~\cite{akkaynak2019sea} (same models and metrics in consideration as Table~\ref{tab_usod}).}
\footnotesize
\begin{tabular}{l||l|l|l|l}
  \Xhline{2\arrayrulewidth}
    \textbf{Model} & Abs Rel & Sq Rel & RMSE & $\log_{10}$  \\ 
    
    \Xhline{2\arrayrulewidth} 
    VGG-UNet (RGB)  & $1.402$ & $0.680$ & $0.439$ & $0.359$  \\
    
    VGG-UNet (R\textbf{M}I) & $1.271$ & $0.563$ & $0.399$ & $0.337$ \\

    ResNet50-UNet (RGB)  & $1.272$ & $0.563$ & $0.397$ & $0.335$ \\
    
    ResNet50-UNet (R\textbf{M}I)  & $1.206$ & $0.613$ & $0.420$ & $0.352$  \\
    
    AdaBins (RGB)  & $1.314$ & $0.663$ & $0.441$ & $0.356$  \\
    
    AdaBins (R\textbf{M}I)  & $1.258$ & $0.617$ & $0.426$ & $0.346$  \\
    
    
    DenseDepth (RGB)  & {$1.073$} & {{$0.448$}} & {{$0.366$}} & {{$0.312$}} \\

    DenseDepth (R\textbf{M}I) & {$1.092$} & {$0.470$} & {$0.370$} & {{$0.312$}} \\
    
    MobileNetV2 (RGB)  & $1.196$ & $0.565$ & $0.408$ & $0.337$ \\
    
    MobileNetV2 (R\textbf{M}I)  & $1.170$ & $0.547$ & $0.400$ & $0.329$ \\
    
    UDepth (RGB)  & $1.304$ & $0.653$ & $0.440$ & $0.355$ \\
    
    
    UDepth (R\textbf{M}I)  & {$1.153$} & {$0.514$} & {$0.388$} & {$0.321$} \\
    \Xhline{2\arrayrulewidth} 
\end{tabular}
\label{tab_seathru}
\vspace{-1mm}
\end{table}

\begin{figure}[h]
\centering
\includegraphics[width=0.96\linewidth]{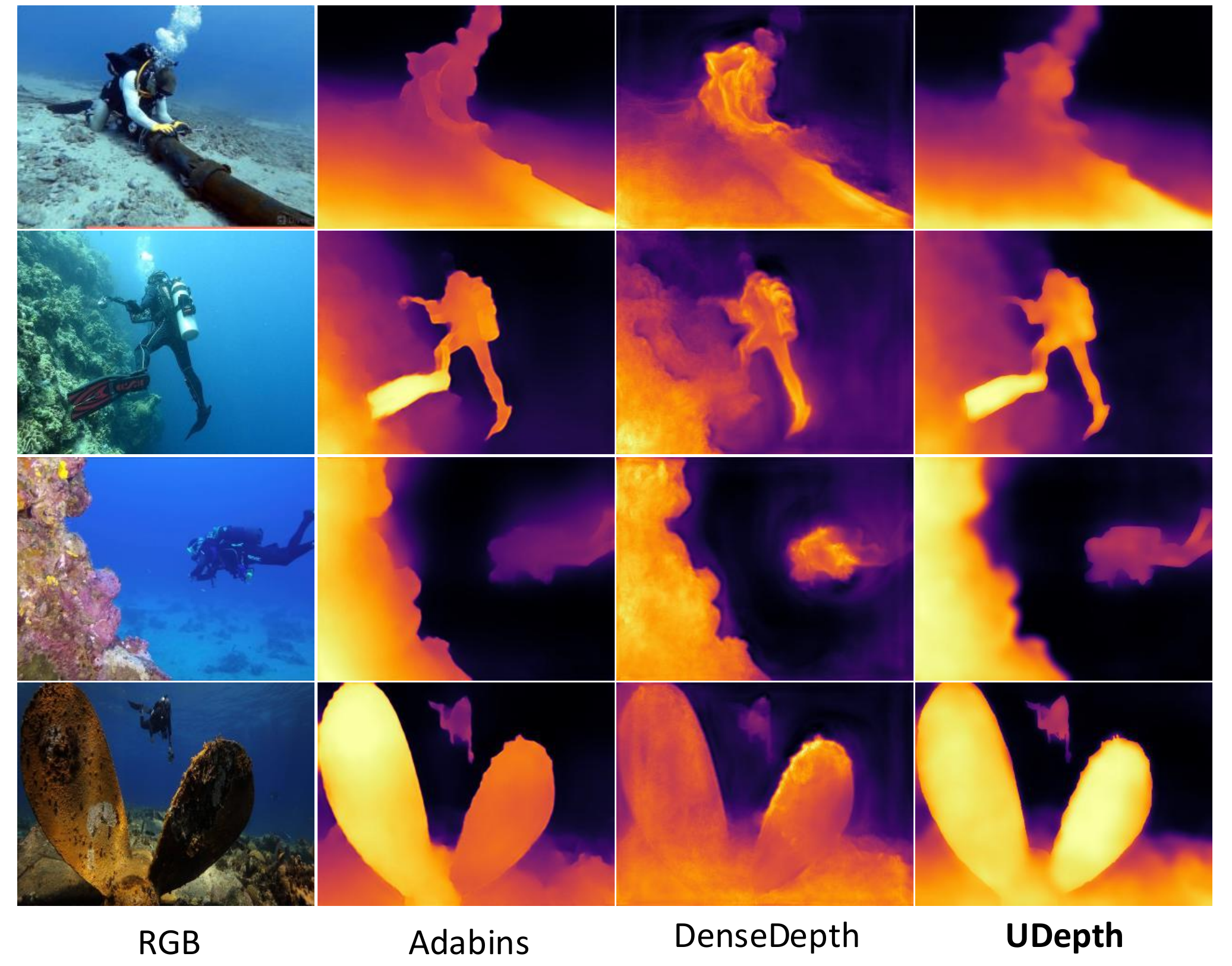}
\vspace{-3mm}
\caption{A few qualitative comparisons for generalization performance of UDepth with its two best competitors: Adabins and DenseDepth are shown. These underwater test images are randomly selected from various benchmark datasets~\cite{islam2022svam,li2019underwater}.}
\label{fig:qual_comp_legacy}
\end{figure}

The qualitative results corroborate our analyses; as the visual comparisons of Fig.~\ref{fig:qual_comp} illustrate, the output of UDepth and Adabins are more accurate and consistent with the underwater scene geometry. Udepth does a particularly better job at removing background regions and predicting foreground layers up to scale. Moreover, UDepth demonstrates much better generalization performance compared to other SOTA models, as evident from the results in Table~\ref{tab_seathru}. Here, we use the D3 (reef) scenes from the Sea-Thru dataset~\cite{akkaynak2019sea} for testing only. While Adabins offers good results, UDepth achieves significantly better results across all metrics on unseen test cases. In comparison, DenseDepth model shows slightly better performance on unseen underwater images; however, with $42.6$M parameters, it is about $3\times$ computationally heavier. UDepth's superior performance, particularly for background segmentation and depth continuity on unseen natural underwater scenes are illustrated in Fig.~\ref{fig:qual_comp_legacy}.

    
    
    
    
    
    
    
    
    
    
    

\begin{table}
\centering
\caption{Comparison of model parameters, model sizes, and inference rates for three best performing models: on a CPU (Intel\texttrademark~Core i9-3.50GHz) and a GPU (NVIDIA\texttrademark~RTX $3080$).}
\footnotesize
\begin{tabular}{l|r||r|r|r}
  \Xhline{2\arrayrulewidth}
    Model & \# Params & Memory & FPS (CPU) & FPS (GPU) \\ \Xhline{2\arrayrulewidth} 
    Adabins & $78.0$ M & $313.8$ MB & $1.98$ & $33.70$ \\ \hline
    DenseDepth & $42.6$ M & $178.3$ MB & $3.06$ & $52.34$ \\ \hline
    UDepth & {$15.6$ M}  & {$62.7$ MB}  & {$13.50$} & {$66.23$} \\ \hline
    \Xhline{2\arrayrulewidth} 
\end{tabular}
\label{tab_comp}
\vspace{-2mm}
\end{table}

\begin{figure}[t]
\centering
\includegraphics[width=0.98\linewidth]{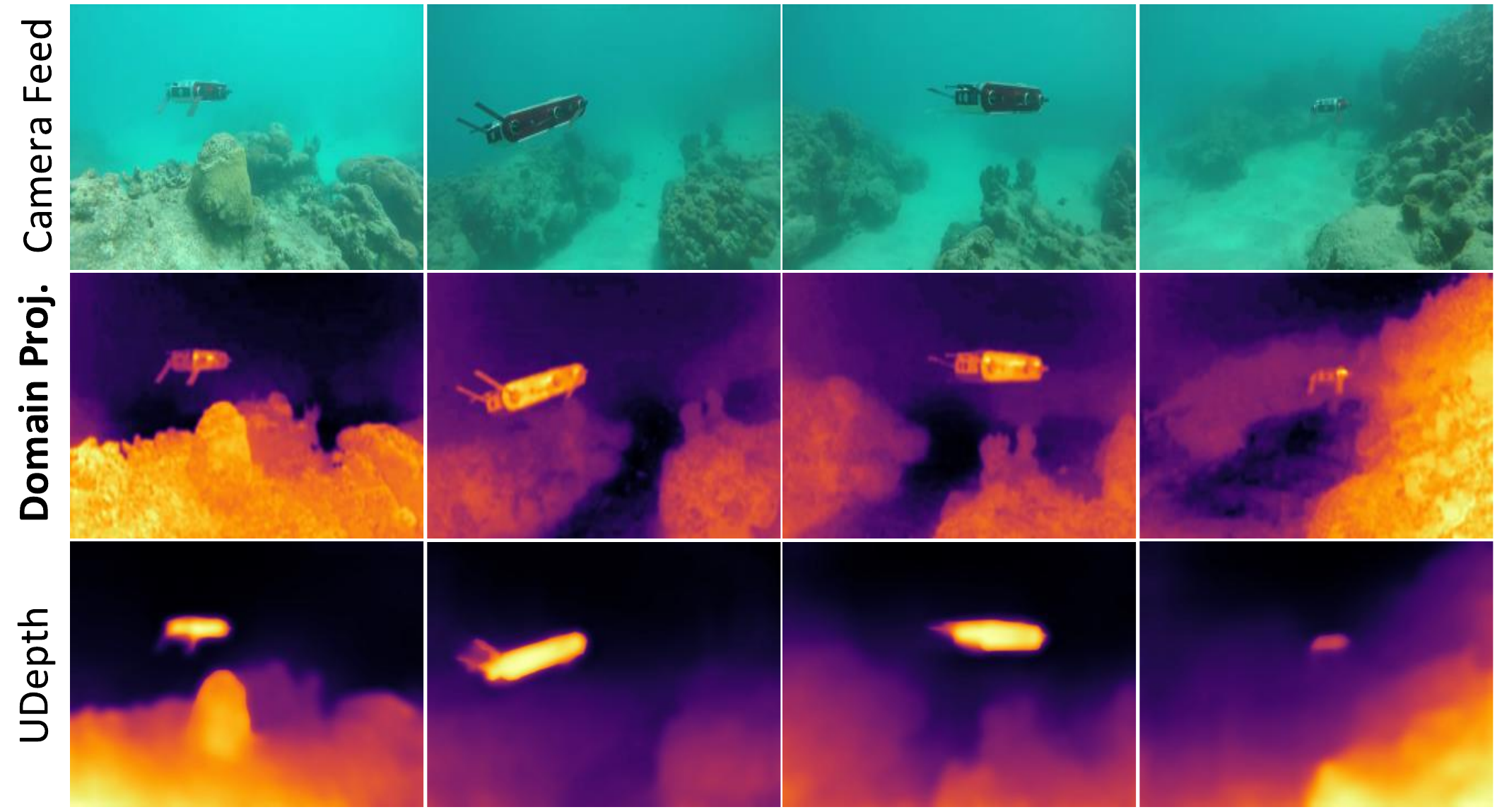}
\vspace{-1mm}
\caption{A few demonstrations of fast coarse depth prediction by our domain projection step are shown in comparison with the final UDepth estimations. The domain projection facilitates an abstract yet reasonably accurate depth prior, which can be used for fast decision-making by underwater robots. }
\label{fig:domain_proj}
\vspace{-6mm}
\end{figure}

\subsection{Coarse Depth Estimation by Domain Projection}
We further compare the computational efficiency of UDepth with Adabins and DenseDepth in Table~\ref{tab_comp}. UDepth is $3$-$5$ times memory efficient and offers $4.4$-$6.8$ times faster inference rates, with over $66$ FPS inference on a single RTX-3080 GPU and over $13.5$ FPS on CPUs. More importantly, we can extend our domain projection step with guided filtering~\cite{he2012guided} for fast \textit{coarse depth prediction} on low-power embedded devices. We performed thorough evaluations on field experimental data, which suggest that these abstract predictions are reasonable approximations of natural underwater scene depths. As shown in Fig.~\ref{fig:domain_proj}, the filtered domain projections embed useful 3D information about the scenes to facilitate high-level decision-making by visually-guided underwater robots. The end-to-end domain projection and filtering module generates depth maps at $51.5$ FPS rate on NVIDIA\texttrademark~Jetson TX2s and on $7.92$ FPS rate on Raspberry Pi-4s.

\vspace{-2mm}
\section{Conclusion}
\vspace{-1mm}
Fast monocular depth estimation can facilitate real-time 3D perception capabilities of autonomous underwater robots. In this work, we propose a deep supervised learning pipeline that includes: ($i$) a domain-aware input space adaptation based on underwater light attenuation characteristics of light propagation; ($ii$) a least-squared formulation of the attenuation constraints for domain projection of coarse underwater scene depth; and ($iii$) an efficient deep visual model named UDepth, which can be trained by that domain projection loss and other pixel-level losses for fine-grained monocular depth estimation. The UDepth model is designed with MobileNetV2 backbone and a Transformer-based optimizer to be able to learn an efficient and robust solution for embedded devices. Experimental results show that with only $20\%$ computational cost, UDepth offers comparable and often better depth estimation performance than SOTA models on benchmark datasets and arbitrary test cases. The domain projection with additional filtering also provides a fast solution for low-powered robots. In the future, we plan to extend UDepth's capabilities toward self-supervised depth estimation and depth-guided image enhancement for real-time underwater robot vision.



{\small
\bibliographystyle{ieee}
\bibliography{main}
}

\end{document}